\documentclass[10pt,conference,letterpaper,final,twocolumn]{IEEEtran}

\usepackage{graphicx}

\begin{document}

\title{An associative memory for the on-line recognition and prediction of temporal sequences}

\author{
\authorblockN{J. Bose\authorrefmark{1}, S.B. Furber\authorrefmark{2}, J.L. Shapiro\authorrefmark{2}}
\authorblockA{School of Computer Science \\
University of Manchester, Manchester M13 9PL, UK}
\authorblockA{\authorrefmark{1}E-mail: joy.bose@cs.manchester.ac.uk}
\authorblockA{\authorrefmark{2}E-mail: [steve.furber, jon.shapiro]@manchester.ac.uk}}

\maketitle

\begin{abstract}   This paper presents the design of an associative memory with feedback that is capable of on-line temporal sequence learning. A framework for on-line sequence
learning has been proposed, and different sequence learning models have been analysed according to this framework. The network model is an associative memory with a separate store
for the sequence context of a symbol. A sparse distributed memory is used to gain scalability. The context store combines the functionality of a neural layer with a shift
register. The sensitivity of the machine to the sequence context is controllable, resulting in different characteristic behaviours. The model can store and predict on-line
sequences of various types and length. Numerical simulations on the model have been carried out to determine its properties. \end{abstract}


\section{Introduction} Many real world problems are sequential in nature, where the time order of events is important. Time can be built into the operation of a neural network
either implicitly, by convolving the weights with samples of the inputs \cite{haykin}, or explicitly, by treating it as an explicit part of the input and thus giving it a spatial
representation. By ``sequence" we refer to a time order of discrete symbols. It is an approximation to temporal coding. A sequence machine is a system capable of storing and
retrieving temporal sequences. 

In this paper we develop a model of on-line sequence learning using Hebbian or one-step learning associative memories. Associative memories have an autonomous learning capability
based on a localised mechanism. The optimal model we used is based on a variant of Kanerva's Sparse Distributed Memory (SDM) that is intrinsically scalable and has been shown to
have good information efficiency \cite{furber}. We have proposed a framework for on-line sequence learning in Section 4, and tried to develop an optimal model based on that
framework. We have also looked at other approaches for sequence learning and the issue of encoding of the past context.

The sequence learning problem can be divided into sequence generation, recognition and prediction \cite{sun}. In this paper we are dealing with the problem of getting the best
prediction after a single presentation of a sequence, not on perfect learning of the patterns generating the sequence after many trials, as dealt with by Schmidhuber
\cite{schmidhuber}, Elman \cite{elman} and others. We treat the whole sequence as one long string without break, and are trying to learn the subsequences in the long unbroken
sequence.

\section{Sequence learning}

\begin{figure}
\centering
\includegraphics[width=2.0in]{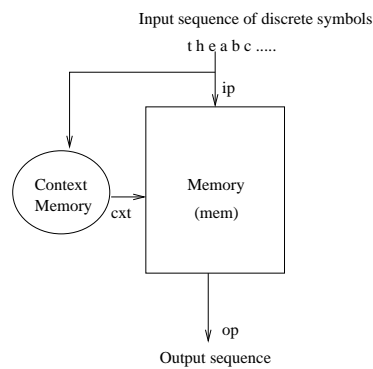}
\caption{Basic design of a sequence memory}
\label{sm_basic}
\end{figure}

Let us say A, B, C are three symbols. Possible sequences can be ABC, BAC etc. Associative memories can learn single associations. We can train such a memory to write the
association A$\to$B, such that when we give A at a later stage (in read mode), we can recover B, even if the input cue we give has been slightly corrupted by noise, e.g. A' rather
than A. 

In building a memory to remember sequences, we need to have some representation of the context of the symbol in the sequence. Two different sequences may have certain symbols in
common, e.g. ABCDE and WXYDZ. If we use an associative memory which can learn only single associations to learn the two above sequences, it cannot decide what is the successor of
symbol D, because the two sequences have different successors of D. After learning the sequences, if, during recall, we give D as input, it cannot decide if the output should be E
or Z on the basis of the present input alone. It needs to have some idea of the context as well. Thus the basic sequence machine needs to have four components: input (ip), output
(op), main memory (mem) and context memory (cxt). Fig.1 gives the design of a basic sequence machine.

Learning in a sequence machine can be on-line or off-line. On-line learning implies that there is no separation between the reading and writing phases. As a sequence is input to the
memory one symbol at a time, the machine calculates the output based on the input and context using the associations formerly written in the memory. This output is a `prediction'
of the next input in the sequence. However if the actual next input is different from the predicted value, it learns to predict the right value next time. In off-line learning, the
reading and writing phases are well separated. The memory learns associations only during the writing phase.

The ideal on-line sequence machine is one that can look back from the most recent inputs as far as is necessary to find a unique context for deciding the next character to be
predicted. The machine should be able to `lock-on' or converge to a context (and thus predict the next output) if it has seen it earlier, and to learn  the new association if it
has not. It should have infinite look-back, yet should be able to distinguish between different contexts. 

\section{Previous approaches and issues with sequence learning} Various approaches to sequence learning in general have been tried by people in the past. Examples are Recurrent
nets with back-propagation, such as by Schmidhuber \cite{schmidhuber}, Hopfield nets (\cite{guyon}), temporal difference and reinforcement learning (\cite{sutton,
worgotter}), hidden Markov models \cite{baum}, self organising  maps \cite{strickert}, competitive  nets\cite{ans} etc. Inspiration from biology has also been used to develop
models of short term memory (\cite{baddeley,fortin,burgess,troyer}).  Sequence learning has been the subject of interest in a variety of other domains as well
(\cite{sun,worgotter}). Sequence machines have been developed for varied tasks such as music composition \cite{mozer1}, protein sequence classification \cite{berry}, robot
movements \cite{ans}, grammar learning \cite{elman}, etc.

Sequence learning models can be classified in various ways, such as the architecture, encoding (linear or non-linear) and decoding (closest match or others) schemes, static (does
not learn) or adaptive memory, representation of the state or history in the memory,  learning algorithm used (correlation or reward based), closed loop (recurrent) or open, etc.
Mozer \cite{mozer} and Sun \cite{sun} have proposed classification schemes and classified many of the existing models in this way. 

Schmidhuber's method is a RNN (recurrent neural net) which can be trained very efficiently and which can remember error signals over long time lags. Reinforcement learning and
gradient methods have been generally favoured in most of the literature, especially where practical applications are concerned. 

The encoding problem is how to represent or  encode the history of the sequence in an effective way so that we can recover the whole sequence in an associative memory, on
presenting this context as a cue. Plate \cite{plate} has dealt with the problem of encoding higher-level associations as a fixed length vector in some detail.

Tino \cite{tino} has dealt with the dynamics of RNN's with random initialised weights (without training). He also developed a Prediction Fractal Machine \cite{tino1}, similar to a
Hidden Markov Model that encoded a sequence as a structure of points on a hypercube and could predict well in one shot.

In the following sections we consider two approaches to deal with our problem (on-line learning using associative memories): delay lines, which store a time window of a fixed
number of last states of the sequence, and neural layer models, which store the state or context or entire history in a neural layer, which is equivalent to a non-linear function
of the past states. We combine the two models mentioned, and show that the combined model performs better than either of them in simulations. We choose this particular approach
over the many others mentioned earlier, partly because of its simplicity, suitability for on-line learning, speed and ease of implementation in associative memories, and partly
because a number of approaches can be thought of as representing one of the above two cases, although in some convoluted or kernel form. 

\section{A framework for on-line sequence learning} When a new symbol is presented at the input of the on-line sequence memory, the memory should learn the association of the
context and the input, and should calculate the output based on this. We can divide the process into the following three steps:

1. The machine associates the new input symbol with the present value of the context. 

2. Based on the new input and the present value of the context, the machine creates a new context. 

3. The machine calculates the output by presenting the modified context to the memory. 

The above steps incorporate both prediction (step 3) and learning (step 1). If the memory has seen a similar input and context before, it will not write anything new to the memory
and the expected next output will be predicted. On the other hand, if it is given a new association, it writes it to the memory. In such a case, the predicted output might be
incorrect, but the memory will learn to give the correct prediction the next time the association is presented.

\section{Modelling issues} In our models, the memory associates the context cue with the input symbol. Both of these are represented as vectors. We use a rank-ordered N-of-M code,
where exactly N of a total of M neurons are active in order to give a valid code, and the firing order of the N active neurons is significant. Thorpe first used rank-ordered codes
for his work \cite{thorpe}.We use N-of-M codes as they are self error-correcting, and ordered codes as they have more information content than unordered binary ones. We represent
the code as a vector where the order is captured by reducing the weight of each successive neuron to fire by a geometric ratio. Thus, for example, the 3-of-5 code representing the
neuron firing order 3-2-4 is represented as [0, k, 1, $k^{2}$, 0] where k$<$1. Each symbol in the input alphabet is given a fixed encoding. We have a real valued
associative memory which learns associations of the context and input vectors. We use the max function (which is a non-linear correlation function) as the training algorithm for
the weights, where the new weight matrix is the maximum of the outer product of the two vectors to be associated and the old weight matrix. For decoding the output, we see which
of the input vectors it is most similar to. Similarity is measured by taking the dot product of the vectors. 

In the model, learning takes place in the main memory only, not in the context memory.

In principle, any associative memory with non-negative real valued or binary weights can be used in the model, but here we used a modified Kanerva Sparse Distributed Memory (SDM)
\cite{kanerva} using rank-ordered N-of-M codes \cite{furber} as the encoding.  The original N-of-M SDM consisted of two layers of neurons: an address decoder layer, whose primary
purpose is to cast the input symbol into high dimensional space to make it more linearly separable, and a second was a correlation-matrix layer \cite{kohenen} called the data
store, which associates the first symbol as decoded by the first neural layer, with the second symbol. Learning takes place only in this layer, while the weights of the first
address decoder layer stay constant. The number of address decoder neurons is much greater than the number of input neurons. Such memories have been proved to be scalable and
error tolerant \cite{furber}. The large size of the address decoder layer is what makes such memories scalable, else they would be identical to correlation matrix memories. 

The operation of the SDM can be described as follows: First of all, the data store gets the input rank-ordered N-of-M encoded word from the encoder, which is one of the  symbols
in the sequence to be stored. The context outputs are fed into the address decoder, whose outputs are fed to the data store too. The data store writes the association  between the
address decoder output and the input data.  After that, the context outputs are fed back to the context layer along with the encoded input, and the context layer generates the new
context. Finally, this new context is fed into the address decoder, whose outputs feed to the data store as before. The data store calculates the final outputs, which are then
decoded. Thus the whole operation proceeds discretely with each input character. 

The sequence machine has three primary components: an encoder to encode the input characters, the neural sequence memory, and a decoder to decode the neural memory output back
into characters. An encoder and decoder translate the input symbols into the desired rank-ordered N-of-M code and back. The characters of the sequence are fed, one at a time, into
the encoder. The encoder converts each character into the appropriate neural code which is input to the neural memory. The decoder decodes the neural memory outputs. The encoder
has a unique encoding for each character in the input alphabet. If implemented in neurons, the encoder can be represented by a single neural layer having fixed weights, such that
the mappings of the characters to the neural code are fixed: it behaves like a lookup table. The decoder is similar to the encoder, except that the inputs and outputs are
reversed. However, the decoder must also have the ability to distinguish between characters with errors and errors that look like characters: it must have a threshold so that
signals that are too weak are not interpreted as characters. The purpose of the decoder is not just to output the closest matching character to the neural code, but also to check
if the code  is sufficiently close to one of the stored characters. 

In the following sections we describe how we dealt with the issue of finding a suitable representation of the context, which encapsulates the past history of the sequence. 
\section{The Shift Register model}

\begin{figure}
\centering
\includegraphics[width=2.0in]{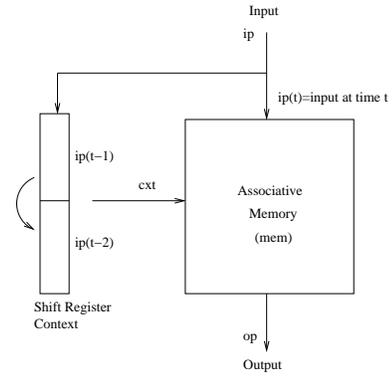}
\caption{Two-Shift register model of a sequence memory}
\label{shiftreg}
\end{figure}

One way to represent the context could be to have a fixed length time window of the past, and associate the next output with inputs in the time window, as is done in Time Delay
Neural Nets (TDNN) \cite{hinton}. Such a memory acts like a shift register. Relating this model to the on-line learning framework described in section 4, here in Step 2 the new
context will be obtained by adding the input to a shifted version of the old context. 

Fig.2 shows the design of a shift register model. An advantage of using a shift register model is that we can retrieve the rest of a stored sequence of any length by  giving any
inputs starting from the middle of a sequence. The disadvantage in using this model is  that the time window is of fixed size, and the number of common symbols might be greater
than the size. The shift register forgets the old context beyond the look-back: for example a 2-shift register can remember the previous two characters at best. Here the context
is a linear function of the previous two inputs. The more recent inputs can be made more important than the old ones by multiplying the context value by a constant $<$1. 

\section{The context neural layer model}

Another common approach is to use a separate `context' neural layer to represent the entire history of the sequence, rather than only a fixed-length time window. This separate
neural layer would store a representation of the context or past history of the sequence, rather than  just the last few symbols as in a shift register model. In such a memory,
when we give an input symbol and want the output according to  the sequences previously learnt by the memory, it is determined by the present input as well as the output of this
context layer, which is a non-linear encoding of all the past inputs. Fig.3 gives the structure of the context layer based model.

\begin{figure}[h!]
\centering
\includegraphics[width=2.0in]{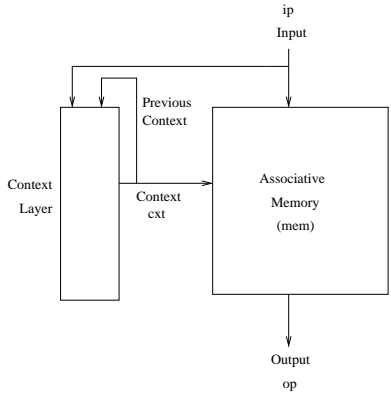}
\caption{Using a separate neural layer as context}
\label{neurallayer}
\end{figure}

Relating this separate context neural layer model to the framework in section 4, here in Step 2 the new context is the output of the context neural layer, whose inputs are the old
 context and the present input symbol. The new context will be the output of the context neural layer with fixed weights whose inputs are the fed-back previous context and the
 input. The influence of the old context can be modulated by multiplying the old context inputs to the neural layer by a constant $\lambda$, the context sensitivity factor. Thus
 the context encodes the entire past history or `state' of the sequence. 

Such a model resembles a finite state machine and was used by Elman \cite{elman} and Jordan \cite{jordan}. Such models can theoretically give unlimited look-back, as the entire
history of the sequence is stored in the memory. However, a problem with the context neural layer model is that to retrieve a sequence we need to start retrieval from the
beginning of the sequence. To solve this, the effect of the context can be modulated by using a fixed modulation factor. This way we can ensure that the past history is  slowly
forgotten, and the present inputs have a greater role than the past in determining what the next outputs should be. This is to ensure that a noisy input symbol in the middle of a
sequence while the memory is in learn mode, or at the  beginning while the memory is in recall mode, does not mess up all the future outputs. 

\section{Combined model based on both context based and shift register model} The shift register model and the separate context layer model both have their advantages and
disadvantages as stated in Sections 6 and 7 respectively. We combine the  two in a new memory model by using a separate context layer with modulated context, where the new context
is determined by both the input  and a shifted version of the present context.  

\begin{figure}[h!]
\centering
\includegraphics[width=2.0in]{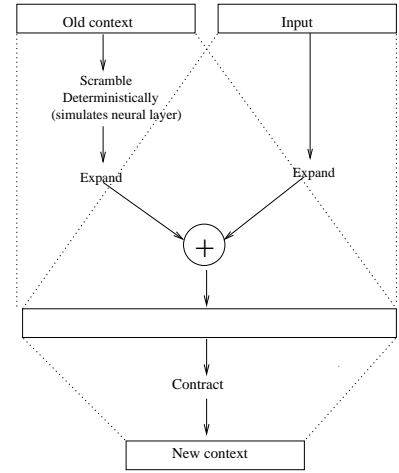}
\caption{Creation of the new context from the old context and the input. The model has aspects of both the neural layer and the shift register}
\label{combinedcontext}
\end{figure}

Figure 4 shows how the new context is formed from the old context and input. Here the context is set to an ordered K-of-M code, where K$>>$N, while the inputs and outputs are
coded as ordered N-of-M.

The context is modified in the following way:

\textbf{Step 1}: The old context is passed through a fixed scrambler, that scrambles it deterministically (representing a neural layer). 

\textbf{Step 2}: The scrambled version of the old context is then multiplied by a scaling factor (x$<$1), which is then added to the rank-ordered N of M input code, and K maximum
components of it are chosen to make the new context. This ensures that the most important bits of the input replace those of the old context, and the old context bits get shifted
to less important bits of the new context. Thus the input bits are shifted up, and context bits down. Thus this part represents a shift register.

\begin{figure}
\centering
\includegraphics[width=2.0in]{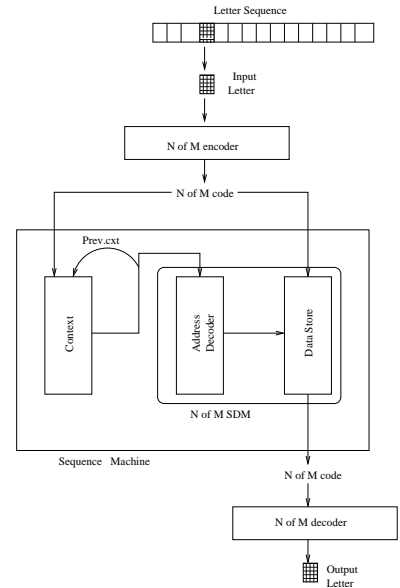}
\caption{A sequence machine using an N of M Kanerva type network for scalability, having address decoder, data memory and context layers}
\label{singleman}
\end{figure}

Figure 5 shows the complete model built in this way.

\section{Numerical tests on the combined sequence machine}
We conducted some tests on the sequence machines described above, to analyse their behaviour with different kinds of sequences. The tests were for on-line learning of symbol 
sequences, as described in the framework in Section 4. The memory was expected to learn previously unseen sequences in a single pass. There are three kinds of sequence machine we
are comparing, namely the shift register, context neural layer and combined model. The basic features of the three models have already been described in previous sections.

The memory used in these experiments had a size of 2048 address decoder neurons, 512 context neurons and 256 data memory neurons. The code used was an 11-of-256 code. Here
the length of the context (512) has been kept double of the input (256), so in the shift register model, we have a look-back of 2. 

\subsection{Comparison of different models. }
In the first experiment we compare the three models of sequence machine and analyse their performance for different sequence lengths. The alphabet size is 15, therefore sequences
of more than 15 characters are bound to include repeats.

\begin{figure}[h!]
\centering
\includegraphics[width=2.5in]{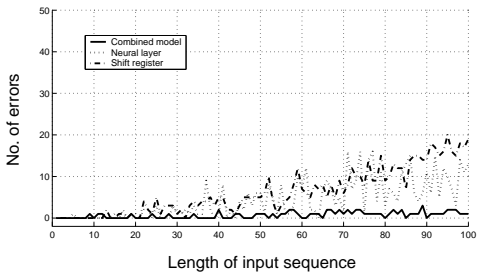}
\caption{Comparison of the performance of three types of sequence memories: Shift register (dashed and dotted line), neural layer(dotted) and the combined model(solid
line). Optimal parameters have been used. The combined model performs better (least number of errors) than the others.}
\label{lambdacompare}
\end{figure} 

Figure 6 shows the results of the first experiment. For each point on the figure we started with a blank memory and input the sequence twice. The memory learns the sequence on the
first presentation of the input, and in the second time we check the predicted output sequence to see how accurate the prediction is. The parameters for the respective models have
been optimised. So $\lambda$ is 0.2 (optimum) in the neural layer and 1.0 in the combined model. We see that the combined model (thick dotted line) performs the best of the three
and obtains near perfect recall. We also found that as the alphabet length gets smaller, the three curves diverge more (the combined model is consistently better), as the three
memories respond differently to more number of repeated characters in the sequence. For an alphabet of 50, the three perform nearly the same, which is close to perfect recall. 

\subsection{Behaviour of the models with repeated subsequences. }
In another experiment, we tried to study the behaviour of the three models when the sequence has a certain number of characters in common.  The sequence is of the type
[seq0][common][seq1][common][seq2] where seq0, seq1 and seq2 are subsequences of different lengths and `common' is the common subsequence. Not surprisingly, the shift register
model, having a look-back of 2, could not discriminate between the two `common' subsequences (whose lengths were greater than 2) and so failed to predict the next characters. 

\begin{table}[h!]
\centering
\begin{tabular}{|l|c|c|}
\hline
Model & Avg over 30 runs & No of Times\\
& (out of 23) & Perfect Recovery\\
\hline
Combined & 22.03 & 10\\
\hline
Neural layer & 20.30 & 4 \\
\hline
Shift Reg & 20.47 & 0\\
\hline
\end{tabular}

\caption{Comparison of the performance of the three sequence memory models, with a repeated subsequence }

 \end{table}

Table 1 shows one such experiment, when a subsequence of length 4 is repeated. The results are averaged over 30 runs. Alphabet length is 12.
The total sequence length is 23 (5-4-5-4-5 according to the description in the previous paragraph). The memory sizes and other optimised parameters are same as in last experiment.
The combined model performs better and gets more perfect recalls than the others, while the shift register never gets perfect recall, as its look-back is smaller than number of
repeated characters

\subsection{Effect of the context sensitivity factor. }
\begin{figure}[h!]
\centering
\includegraphics[width=2.5in]{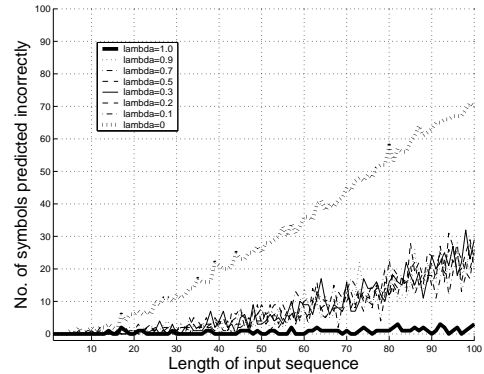}
\caption{Performance of the combined model with context sensitivity $\lambda$ in an experiment to see the effect of  $\lambda$ on memory performance. The highest dotted curve
is for $\lambda$ 0 and lowest solid thick curve for $\lambda$ 1.0}
\label{lambdavary}
\end{figure} 


In another experiment, shown in Fig.7, we vary the context sensitivity factor $\lambda$ to see the memory performance. We see three clear zones. One, when $\lambda$ is 0, the
machine is not at all sensitive to context and it performs badly. When it is 1, which means that the context is given equal priority as the current input, it performs quite well,
with very few errors. When it is between these two zones, the combined model effectively behaves like a shift register. 

\subsection{Use of repeated training (as opposed to one shot learning). } We found that repeated training did not have much use on the machine in most cases, and a single pass gave
fairly good results, since the memory is such that any number of writings to the memory is same as one writing, in cases where the interference noise is not too high. However, in
such cases where it predicted incorrectly, the memory performance improved on repeated training of the same sequence. In the context layer model, the contexts are different each
time, but they converge soon and repeated writing still has no effect. Also, the machine has problems un-learning previously learnt associations, as what is written to the memory
cannot be erased.

\section{Conclusions and further work} We have thus developed a neural network model that is capable of on-line learning and recall of symbol sequences. We are currently
investigating the possibilities of such memories being implemented by real time asynchronous spiking neurons. Work also needs to be done to develop suitable applications where the
model can be used. 

We have built a context-based associative memory in which the influence of the context can be dynamically tuned. Through experiments we have measured the performance of the memory
in storing large sequences of symbols and recalling them successfully. 

\section{Acknowledgement}
J. Bose gratefully acknowledges support from ORS studentship, courtesy of Universities UK.

\bibliographystyle{ieeetr}

\end{document}